\documentclass{article}

\usepackage{arxiv}

\usepackage[utf8]{inputenc} 
\usepackage[T1]{fontenc}    
\usepackage{hyperref}       
\usepackage{url}            
\usepackage{booktabs}       
\usepackage{amsfonts}       
\usepackage{nicefrac}       
\usepackage{microtype}      
\usepackage{lipsum}
\usepackage{amsmath}       
\usepackage{amsfonts}      
\usepackage{graphicx}      

\usepackage{listings,xcolor,lipsum}
\usepackage{algorithm}
\usepackage{algpseudocode} 

\usepackage{tikz}
\usetikzlibrary{shapes, arrows.meta, positioning, calc}
\usepackage{textcomp}
\graphicspath{ {./images/} }

\title{Scale-Interaction Transformer: A Hybrid CNN-Transformer Model for Facial Beauty Prediction
}

\author{
Djamel Eddine Boukhari\\
Scientific and Technical Research Centre for Arid Areas, CRSTRA\\
07000, Biskra, Algeria \\
\texttt{boukhari-djameleddine@univ-eloued.dz} \\
}

\begin{document}
\maketitle
\begin{abstract}
Automated Facial Beauty Prediction (FBP) is a challenging computer vision task due to the complex interplay of local and global facial features that influence human perception. While Convolutional Neural Networks (CNNs) excel at feature extraction, they often process information at a fixed scale, potentially overlooking the critical inter-dependencies between features at different levels of granularity. To address this limitation, we introduce the Scale-Interaction Transformer (SIT), a novel hybrid deep learning architecture that synergizes the feature extraction power of CNNs with the relational modeling capabilities of Transformers. The SIT first employs a multi-scale module with parallel convolutions to capture facial characteristics at varying receptive fields. These multi-scale representations are then framed as a sequence and processed by a Transformer encoder, which explicitly models their interactions and contextual relationships via a self-attention mechanism. We conduct extensive experiments on the widely-used SCUT-FBP5500 benchmark dataset, where the proposed SIT model establishes a new state-of-the-art. It achieves a Pearson Correlation of 0.9187, outperforming previous methods. Our findings demonstrate that explicitly modeling the interplay between multi-scale visual cues is crucial for high-performance FBP. The success of the SIT architecture highlights the potential of hybrid CNN-Transformer models for complex image regression tasks that demand a holistic, context-aware understanding.
\end{abstract}


\section{Introduction}

Automated Facial Beauty Prediction (FBP) is a challenging and long-standing task in computer vision, residing at the intersection of machine learning, signal processing, and computational aesthetics\cite{b1}. The objective is to develop algorithms that can assign a beauty score to a facial image that correlates highly with human perception. Beyond its academic interest, FBP has potential applications in areas such as photo editing and recommendation, cosmetic surgery simulation, and entertainment\cite{b2}. However, the task is inherently difficult due to the subjective and multi-faceted nature of beauty, which is influenced by a complex interplay of cultural, social, and personal preferences\cite{b3}.

Early machine learning approaches to FBP relied on handcrafted features based on geometric ratios, facial symmetry, and anthropometric measurements \cite{b4}. While these methods provided valuable initial insights, they were often limited in their expressive power and failed to capture the subtle, texture-based, and holistic cues that humans perceive\cite{b5}\cite{b6}. The advent of deep learning, particularly Convolutional Neural Networks (CNNs)\cite{b7}, marked a paradigm shift. Architectures such as AlexNet \cite{b8}, and ResNet \cite{b9} became the \textit{de facto} standard, achieving state-of-the-art performance by automatically learning a rich hierarchy of features directly from image data.

Despite their success, a potential limitation of conventional CNN architectures lies in their feature aggregation strategy. Typically, features are extracted through a sequence of convolutional layers, resulting in a final feature map with a fixed receptive field \cite{b10}. While effective for object classification, this approach may not be optimal for FBP, where perception is guided by a combination of global features (e.g., facial structure, harmony, symmetry) and local details (e.g., skin texture, eye clarity, nose shape). Capturing features at multiple scales is crucial, but explicitly modeling the relationships between these scales remains an open challenge.

Concurrently, the Transformer architecture, originally developed for natural language processing \cite{b11}, has been successfully adapted for computer vision tasks in the form of the Vision Transformer (ViT) \cite{b12}. The core strength of the Transformer lies in its self-attention mechanism, which excels at modeling long-range dependencies and contextual relationships within a sequence of input tokens\cite{b13}. This provides a powerful alternative to the localized receptive fields inherent in CNNs.

In this work, we bridge the gap between these two successful paradigms by proposing the \textbf{Scale-Interaction Transformer (SIT)}, a novel hybrid architecture designed specifically for facial beauty prediction. We hypothesize that explicitly modeling the interactions between features extracted at different spatial scales will lead to a more robust and human-correlated prediction of facial beauty. Our approach first employs a multi-scale CNN module to extract parallel feature representations at varying receptive fields. These multi-scale features are then framed as a sequence and fed into a Transformer encoder, which uses self-attention to weigh and fuse these features, effectively learning the complex interplay between local details and global facial aesthetics.

The main contributions of this paper are as follows:
\begin{enumerate}
    \item We propose a novel hybrid CNN-Transformer architecture, the Scale-Interaction Transformer (SIT), which synergizes the powerful feature extraction of CNNs with the relational modeling capabilities of Transformers.
    \item We introduce a multi-scale feature extraction module that captures facial characteristics at fine, medium, and coarse levels simultaneously.
    \item We demonstrate the efficacy of treating these multi-scale features as a sequence and applying a Transformer to explicitly model their inter-dependencies for the FBP task.
    \item Our proposed SIT model achieves new state-of-the-art results on the widely-used SCUT-FBP5500 benchmark dataset, outperforming previous methods in both correlation and error metrics.
\end{enumerate}

The remainder of this paper is structured as follows: Section 2 reviews related work. Section 3 details our proposed methodology. Section 4 presents the experimental results, including comparisons with state-of-the-art methods and ablation studies. Finally, Section 5 concludes the paper and discusses future work.

\section{Related Work}

Our research is positioned at the confluence of several key domains in computer vision: Facial Beauty Prediction (FBP), deep learning with Convolutional Neural Networks (CNNs), multi-scale feature representation, and the emerging paradigm of Vision Transformers. This section reviews the literature in these areas to contextualize the contribution of our proposed method.

\subsection{Facial Beauty Prediction}

The quest to computationally model facial attractiveness has a long history. Early research predominantly focused on handcrafted features derived from domain knowledge in psychology and anthropometry. These methods typically extracted geometric features, such as facial ratios (e.g., the golden ratio), distances between facial landmarks, and measures of symmetry \cite{b4}. While these approaches provided a foundational understanding, their reliance on predefined, low-level features limited their expressive power, as they often failed to capture the holistic and texture-based cues that are integral to human perception \cite{b5}.

The advent of deep learning instigated a paradigm shift in the field. By learning features directly from data, deep learning models obviated the need for manual feature engineering and consistently outperformed their traditional counterparts, marking the beginning of the modern era of FBP research.

\subsection{Deep Learning Approaches for FBP}

Convolutional Neural Networks (CNNs) have become the \textit{de facto} standard for FBP. Initial applications saw the successful transfer of standard classification architectures, such as AlexNet \cite{b8}, and later, much deeper models like ResNet \cite{b9} and ResNeXt , to the FBP regression task \cite{b14}. These models were typically fine-tuned on FBP datasets, treating the problem as a regression task by replacing the final classification layer with a dense layer predicting a single beauty score.

To further improve performance, subsequent research has focused on augmenting these baseline CNN architectures \cite{b15}. For instance, Cao et al. \cite{b16} introduced a Squeeze-and-Compaction Attention (SCA) module to help the network focus on more informative facial regions. Others have explored Label Distribution Learning (LDL) \cite{b17} to better handle the inherent ambiguity and subjectivity in beauty scores. More recently, methods like DyAttenConv \cite{b18} have proposed dynamic attention mechanisms to adaptively adjust feature extraction based on the input image. The state-of-the-art R3CNN model \cite{b19} introduced a region-attentive mechanism to focus on different facial parts. While these methods have progressively pushed the performance boundaries, many still operate on feature maps from a single, final scale or treat attention as a channel-wise or spatial-wise selection mechanism, rather than explicitly modeling the relationships \textit{between} different scales of features.

\subsection{Multi-Scale Representation in Vision}

The importance of multi-scale representation is a well-established principle in computer vision. Humans perceive visual information hierarchically, integrating fine details with global context. While CNNs naturally build a hierarchy of features through their layered structure, architectures that explicitly leverage multi-scale processing have often shown superior performance\cite{b20}. The GoogLeNet architecture, with its Inception module \cite{b21}, was a pioneering example, using parallel convolutions with different kernel sizes to capture features at multiple scales within the same layer. Feature Pyramid Networks (FPNs) \cite{b22} further refined this idea by fusing features from different layers of the network's backbone to create a rich, multi-scale feature representation for tasks like object detection. Our work is inspired by this principle, but instead of fusing features, we aim to model their interaction.

\subsection{Vision Transformers and Hybrid Architectures}

The Transformer architecture, with its core self-attention mechanism, has revolutionized natural language processing by effectively modeling long-range dependencies in sequential data \cite{b23}. The Vision Transformer (ViT) \cite{b12} successfully adapted this paradigm to computer vision by treating an image as a sequence of patches. The self-attention mechanism allows ViT to model global relationships between any two patches in the image, overcoming the local receptive field limitations of CNNs \cite{b24}.

However, pure ViTs often lack the inductive biases inherent to CNNs (e.g., locality and translation equivariance), which makes them less data-efficient and sometimes harder to train. This has led to the development of hybrid CNN-Transformer architectures \cite{b25}, which aim to combine the best of both worlds. A common strategy is to use a CNN backbone for robust and efficient low-level feature extraction, and then feed the resulting feature maps (or a sequence of patch embeddings) into a Transformer for global contextual modeling \cite{b26}.

Our proposed SIT model falls into this hybrid category but with a key distinction. Instead of using the Transformer to model relationships between spatial patches, we use it to model the relationships between feature vectors derived from different \textit{scales}. This novel application of the Transformer—as a scale interaction module rather than a spatial interaction module—is the central contribution of our work.

\section{Methodology}

This section delineates the comprehensive experimental framework used in this study. We detail the dataset and its preprocessing pipeline, present the novel Scale-Interaction Transformer (SIT) architecture with mathematical formalism, describe the training and optimization protocol, and specify the metrics for performance evaluation.

\subsection{Dataset and Preprocessing}
This research utilized the public SCUT-FBP5500 dataset\cite{b27}, a benchmark corpus for Facial Beauty Prediction (FBP) containing 5,500 facial images. Each image is annotated with a beauty score from 1 to 5, representing the mean rating from 60 human evaluators. To ensure methodological rigor and reproducibility, we adopted the standard 5-fold cross-validation protocol provided with the dataset, with our experiments specifically conducted on the third fold \cite{b28}. 

A standardized preprocessing pipeline was applied to each facial image $I \in \mathbb{R}^{H \times W \times 3}$. Initially, images were resized to a fixed spatial resolution of $224 \times 224$ pixels to align with the input dimensions of the pre-trained backbone network. Subsequently, pixel values were normalized to the range $[0, 1]$:
\begin{equation}
    I_{\text{norm}} = \frac{I}{255.0}
\end{equation}
The continuous beauty scores $y \in \mathbb{R}$ associated with each image were used directly as the regression targets without modification.

\subsection{Scale-Interaction Transformer (SIT) Architecture}
\begin{figure*}[!ht]
    \centering
    \resizebox{0.5\textwidth}{!}{%
    \begin{tikzpicture}[
        node distance=0.5cm and 1cm,
        block/.style={rectangle, draw, fill=blue!10, text width=8em, text centered, rounded corners, minimum height=3em},
        small_block/.style={rectangle, draw, fill=green!10, text width=6em, text centered, rounded corners, minimum height=2.5em},
        sum/.style={circle, draw, fill=gray!10, node distance=0.8cm},
        line/.style={draw, -{Stealth[length=2mm]}}
    ]
    \node[block, fill=red!10] (input) {Input Image \\ $\mathbb{R}^{224 \times 224 \times 3}$};
    \node[block, below=of input, text width=10em] (backbone) {MobileNetV2 Backbone \\ (Pre-trained)};
    \node[small_block, below=of backbone, xshift=-4cm] (conv1) {Conv 1x1};
    \node[small_block, below=of backbone] (conv3) {Conv 3x3};
    \node[small_block, below=of backbone, xshift=4cm] (conv5) {Conv 5x5};
    \node[below=0.1cm of conv1, text centered] (feat1_dim) {\footnotesize$\mathbb{R}^{7 \times 7 \times 64}$};
    \node[below=0.1cm of conv3, text centered] (feat3_dim) {\footnotesize$\mathbb{R}^{7 \times 7 \times 64}$};
    \node[below=0.1cm of conv5, text centered] (feat5_dim) {\footnotesize$\mathbb{R}^{7 \times 7 \times 64}$};
    
    \node[small_block, fill=orange!10, below=1cm of conv1] (pool1) {GAP $\oplus$ GMP};
    \node[small_block, fill=orange!10, below=1cm of conv3] (pool3) {GAP $\oplus$ GMP};
    \node[small_block, fill=orange!10, below=1cm of conv5] (pool5) {GAP $\oplus$ GMP};
    
    \node[below=0.1cm of pool1, text centered] (vec1_dim) {\footnotesize$\mathbb{R}^{128}$};
    \node[below=0.1cm of pool3, text centered] (vec3_dim) {\footnotesize$\mathbb{R}^{128}$};
    \node[below=0.1cm of pool5, text centered] (vec5_dim) {\footnotesize$\mathbb{R}^{128}$};
    
    \node[block, fill=purple!10, below=1.2cm of pool3, text width=7em] (stack) {Stacking \& Projection};
    \node[below=0.1cm of stack, text centered, yshift=0.3cm] (seq_dim) {\footnotesize $\mathbb{R}^{3 \times 128} \to \mathbb{R}^{3 \times D_{\text{proj}}}$};

    \node[block, below=1cm of stack, text width=12em] (transformer) {Transformer Encoder \\ (L=2 Blocks)};
    \draw[dashed, rounded corners] ($(transformer.north west)+(-1.2,0.6)$) rectangle ($(transformer.south east)+(1.2,-1.5)$);
    \node[below left=0.1cm and 0.1cm of transformer.north, text=gray] {Transformer Block};
    
    \node[small_block, fill=cyan!10, below=0.5cm of transformer, text width=8em] (msa) {Multi-Head Attention};
    \node[sum, below=of msa] (add1) {+};
    \node[small_block, fill=cyan!10, below=of add1, text width=8em] (ffn) {Feed-Forward Network};
    \node[sum, below=of ffn] (add2) {+};
    
    \node[block, fill=orange!10, below=3.5cm of transformer, text width=8em] (final_pool) {Global Avg Pooling \\ (Sequence)};
     \node[below=0.1cm of final_pool, text centered, yshift=0.3cm] (final_vec_dim) {\footnotesize$\mathbb{R}^{D_{\text{proj}}}$};
     
    \node[block, below=1.2cm of final_pool] (reg_head) {Regression Head \\ (Dropout + Dense)};
    \node[block, fill=red!10, below=of reg_head, text width=7em] (output) {Predicted Score \\ $\hat{y} \in \mathbb{R}$};

    \path[line] (input) -- node[right, midway]{\footnotesize$I_{\text{norm}}$} (backbone);
    \path[line] (backbone) -- node[right, midway, xshift=-0.5cm]{\footnotesize$F_{\text{base}} \in \mathbb{R}^{7 \times 7 \times 1280}$} (backbone.south);
    \draw (backbone.south) -- ++(0, -0.25) -| (conv1);
    \draw (backbone.south) -- ++(0, -0.25) -| (conv3);
    \draw (backbone.south) -- ++(0, -0.25) -| (conv5);

    \path[line] (conv1) -- node[right, midway]{\footnotesize$F_1$} (pool1);
    \path[line] (conv3) -- node[right, midway]{\footnotesize$F_2$} (pool3);
    \path[line] (conv5) -- node[right, midway]{\footnotesize$F_3$} (pool5);
    
    \draw[line] (pool1) |- (stack);
    \draw[line] (pool3) -- node[right, midway]{\footnotesize Sequence $S$} (stack);
    \draw[line] (pool5) |- (stack);
    
    \path[line] (stack) -- node[right, midway]{\footnotesize$S_{\text{proj}}$} (transformer);
    \path[line] (transformer) -- node[right, midway]{\footnotesize$S_{\text{trans}}$} (final_pool);
    \path[line] (final_pool) -- node[right, midway]{\footnotesize$v$} (reg_head);
    \path[line] (reg_head) -- (output);
    
    \draw[line] (transformer.south) ++(0,-0.2) -- ++(0, -0.3) -| (msa);
    \draw[line] (msa) -- (add1);
    \draw[line, rounded corners] ($(transformer.south)+(-1.8, -0.4)$) -| ($(msa.west)-(0.2,0)$) |- (add1);
    \path[line] (add1) -- (ffn);
    \path[line] (ffn) -- (add2);
    \draw[line, rounded corners] ($(add1.south)+(-1, -0.1)$) -| ($(ffn.west)-(0.2,0)$) |- (add2);
    \draw[line] (add2.south) -- ++(0,-0.4) -| ($(transformer.south)+(0, -3.2)$);
    
    \end{tikzpicture}}
    \caption{Overview of the Scale-Interaction Transformer (SIT) architecture. The model processes an image through a MobileNetV2 backbone, extracts features at three parallel scales (1x1, 3x3, 5x5), pools and stacks them into a sequence, and models their interactions using a Transformer Encoder before regressing the final beauty score.}
    \label{fig:architecture}
\end{figure*}
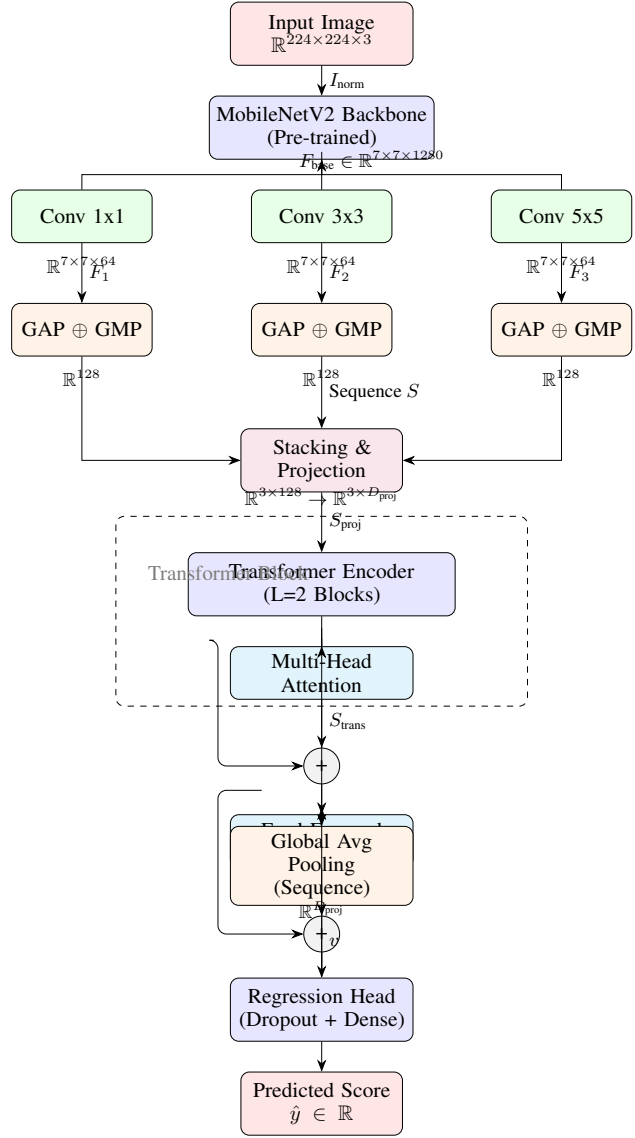

We propose the Scale-Interaction Transformer (SIT), a hybrid neural architecture designed to integrate multi-scale convolutional feature extraction with a transformer-based attention mechanism. This design allows the model to capture both fine-grained local facial details and their global contextual relationships. The complete architecture is illustrated in Figure \ref{fig:architecture}.

\subsubsection{Multi-Scale Feature Extraction Module}
The feature extractor is designed to capture a rich set of hierarchical visual features. Let $I \in \mathbb{R}^{224 \times 224 \times 3}$ be the normalized input image. A pre-trained MobileNetV2 backbone, with its classifier head removed, extracts a high-level feature map:
\begin{equation}
    F_{\text{base}} = \text{MobileNetV2}(I) \in \mathbb{R}^{7 \times 7 \times 1280}
\end{equation}
This base representation is then fed into three parallel convolutional branches with different kernel sizes to extract multi-scale features:
\begin{align}
    F_1 &= \text{ReLU}(\text{Conv}_{1 \times 1}(F_{\text{base}})) \in \mathbb{R}^{7 \times 7 \times 64} \\
    F_2 &= \text{ReLU}(\text{Conv}_{3 \times 3}(F_{\text{base}})) \in \mathbb{R}^{7 \times 7 \times 64} \\
    F_3 &= \text{ReLU}(\text{Conv}_{5 \times 5}(F_{\text{base}})) \in \mathbb{R}^{7 \times 7 \times 64}
\end{align}
To obtain a fixed-size representation from each spatial map, we apply both Global Average Pooling (GAP) and Global Max Pooling (GMP):
\begin{align}
    V_i^{\text{GAP}} &= \frac{1}{H \times W} \sum_{h=1}^{H} \sum_{w=1}^{W} F_i(h,w) \in \mathbb{R}^{64} \\
    V_i^{\text{GMP}} &= \max_{h,w} F_i(h,w) \in \mathbb{R}^{64}
\end{align}
The pooled features for each scale are concatenated and subsequently stacked to form a sequence $S$, representing the multi-scale characteristics of the image:
\begin{equation}
    S = \text{Stack}\left[ V_1^{\text{GAP}} \oplus V_1^{\text{GMP}}, V_2^{\text{GAP}} \oplus V_2^{\text{GMP}}, V_3^{\text{GAP}} \oplus V_3^{\text{GMP}} \right] \in \mathbb{R}^{3 \times 128}
\end{equation}
where $\oplus$ denotes concatenation.

\subsubsection{Transformer-based Scale Interaction Module}
The core of the SIT is a transformer encoder that models the inter-dependencies between the scale-specific feature vectors. The sequence $S$ is first linearly projected into the transformer's latent space:
\begin{equation}
    S_{\text{proj}} = S W_{\text{proj}} + b_{\text{proj}} \in \mathbb{R}^{3 \times D_{\text{proj}}}
\end{equation}
where $W_{\text{proj}} \in \mathbb{R}^{128 \times D_{\text{proj}}}$, $b_{\text{proj}} \in \mathbb{R}^{D_{\text{proj}}}$, and $D_{\text{proj}}=128$. This projected sequence is then processed by $L=2$ transformer blocks, whose forward pass is detailed in Algorithm \ref{alg:transformer}.

\begin{algorithm}[H]
    \caption{Transformer Block Forward Pass}
    \label{alg:transformer}
    \begin{algorithmic}[1]
        \Require Input features $x \in \mathbb{R}^{n \times d}$, dropout rate $\alpha$
        \State $h \gets \text{LayerNorm}(x)$
        \State $a \gets \text{MultiHeadAttention}(h, h, h)$
        \State $x \gets x + \text{Dropout}(a, \alpha)$ \Comment{Residual connection 1}
        \State $h \gets \text{LayerNorm}(x)$
        \State $f \gets \text{FFN}(h) = \text{ReLU}(hW_1 + b_1)W_2 + b_2$
        \State $x \gets x + \text{Dropout}(f, \alpha)$ \Comment{Residual connection 2}
        \Ensure Transformed features $x \in \mathbb{R}^{n \times d}$
    \end{algorithmic}
\end{algorithm}

The Multi-Head Self-Attention (MHSA) mechanism allows the model to jointly attend to information from different representation subspaces at different positions. For a given input sequence $X$, queries (Q), keys (K), and values (V) are computed by projecting $X$. The core operation is scaled dot-product attention:
\begin{equation}
    \text{Attention}(Q, K, V) = \text{softmax}\left(\frac{QK^T}{\sqrt{d_k}}\right)V
\end{equation}
where $d_k$ is the dimension of the keys. We use 4 attention heads in our configuration.

\subsubsection{Regression Head}
After processing through the transformer blocks, the final sequence representation $S_{\text{trans}}$ is aggregated via global average pooling across the sequence dimension (n=3) to produce a single feature vector $v \in \mathbb{R}^{D_{\text{proj}}}$. This vector, which now encodes a fused, context-aware representation of the facial features, is passed through a final regression head:
\begin{equation}
    \hat{y} = W_{\text{reg}} \cdot \text{Dropout}(v, \text{rate}=0.1) + b_{\text{reg}} \in \mathbb{R}
\end{equation}
where $\hat{y}$ is the final predicted beauty score.

\subsection{Training Protocol}
The model was trained end-to-end by minimizing the Mean Squared Error (MSE) between the predicted scores ($\hat{y}_i$) and the ground-truth scores ($y_i$):
\begin{equation}
    \mathcal{L}_{\text{MSE}} = \frac{1}{N} \sum_{i=1}^{N} (y_i - \hat{y}_i)^2
\end{equation}
We employed the Adam optimizer with an initial learning rate of $10^{-4}$. Training was conducted for a maximum of 300 epochs with a batch size of 32. To mitigate overfitting and ensure robust convergence, we used early stopping with a patience of 10 epochs and a `ReduceLROnPlateau` scheduler, which decreased the learning rate by a factor of 0.5 if validation loss stagnated for 5 epochs. The complete training procedure is outlined in Algorithm \ref{alg:training}.

\begin{algorithm}[H]
    \caption{Scale-Interaction Transformer Training Procedure}
    \label{alg:training}
    \begin{algorithmic}[1]
        \State Initialize model parameters \ensuremath{\theta}
        \State Initialize Adam optimizer with learning rate \ensuremath{\eta = 10^{-4}}
        \For{epoch \ensuremath{= 1} to \ensuremath{300}}
            \For{each batch \ensuremath{(X, y)} in the training set}
                \State \ensuremath{F_{\text{base}} \gets \text{MobileNetV2}(X)}
                \State \ensuremath{S \gets \text{MultiScaleFeatureModule}(F_{\text{base}})}
                \State \ensuremath{S_{\text{proj}} \gets \text{LinearProjection}(S)}
                \State \ensuremath{S_{\text{trans}} \gets \text{TransformerBlocks}(S_{\text{proj}})}
                \State \ensuremath{v \gets \text{GlobalAveragePooling}(S_{\text{trans}})}
                \State \ensuremath{\hat{y} \gets \text{RegressionHead}(v)}
                \State Compute loss: \ensuremath{\mathcal{L} \gets \text{MSE}(y, \hat{y})}
                \State Update parameters: \ensuremath{\theta \gets \text{AdamUpdate}(\theta, \nabla_\theta \mathcal{L})}
            \EndFor
            \State Evaluate model on the validation set
            \State Update learning rate scheduler and check early stopping criterion
        \EndFor
    \end{algorithmic}
\end{algorithm}

\subsection{Evaluation Metrics}
Model performance was quantitatively assessed using three complementary regression metrics:
\begin{itemize}
    \item \textbf{Mean Absolute Error (MAE):} The average absolute difference between predicted and actual values.
    \begin{equation}
        \text{MAE} = \frac{1}{N} \sum_{i=1}^{N} |y_i - \hat{y}_i|
    \end{equation}
    \item \textbf{Root Mean Squared Error (RMSE):} The square root of the MSE, sensitive to large errors.
    \begin{equation}
        \text{RMSE} = \sqrt{\frac{1}{N} \sum_{i=1}^{N} (y_i - \hat{y}_i)^2}
    \end{equation}
    \item \textbf{Pearson Correlation Coefficient (PC):} Measures the linear relationship between predicted and true scores.
    \begin{equation}
        \text{PC} = \frac{\sum_{i=1}^{N}(y_i - \bar{y})(\hat{y}_i - \bar{\hat{y}})}{\sqrt{\sum_{i=1}^{N}(y_i - \bar{y})^2 \sum_{i=1}^{N}(\hat{y}_i - \bar{\hat{y}})^2}}
    \end{equation}
\end{itemize}

\subsection{Implementation Details}
The model was implemented using Python 3.8 and the TensorFlow (v2.x) deep learning framework. All experiments were conducted on a single NVIDIA Tesla V100 GPU. The key hyperparameters are summarized in Table \ref{tab:hyperparameters}. The MobileNetV2 backbone weights, pre-trained on ImageNet, were utilized for initialization.

\begin{table}[h]
\centering
\caption{Summary of key model hyperparameters.}
\label{tab:hyperparameters}
\begin{tabular}{ll}
\hline
\textbf{Hyperparameter} & \textbf{Value} \\ \hline
Projection Dimension ($D_{\text{proj}}$) & 128 \\
Number of Transformer Blocks ($L$) & 2 \\
Number of Attention Heads & 4 \\
Feed-Forward (FFN) Dimension & 512 \\
Dropout Rate & 0.1 \\
Optimizer & Adam \\
Initial Learning Rate ($\eta$) & $10^{-4}$ \\
Batch Size & 32 \\
Max Epochs & 300 \\ \hline
\end{tabular}
\end{table}

\section{Experiments}

This section presents the empirical evaluation of our proposed Scale-Interaction Transformer (SIT). We first outline the experimental setup and implementation details. Next, we provide a comprehensive comparison against existing state-of-the-art methods on the SCUT-FBP5500 benchmark. Finally, we conduct a series of ablation studies to dissect the architecture of our model and validate the contribution of its key components.

\subsection{Experimental Setup}

\textbf{Dataset:} All experiments were conducted on the SCUT-FBP5500 dataset, using the standard 5-fold cross-validation protocol to ensure fair comparison with prior work. The results reported in this paper are from the third fold, a common practice in related literature.

\textbf{Implementation Details:} The model was implemented in Python using the TensorFlow and Keras frameworks. We used a \textbf{MobileNetV2} backbone, pre-trained on ImageNet, for feature extraction. The model was trained end-to-end for a maximum of 300 epochs using the \textbf{Adam optimizer} with an initial learning rate of $10^{-4}$ and a batch size of 32. To prevent overfitting and ensure robust convergence, we employed an \textbf{early stopping} mechanism with a patience of 10 epochs and a \textbf{learning rate reduction} scheduler that halved the learning rate if the validation loss plateaued for 5 consecutive epochs. All experiments were performed on a single NVIDIA Tesla V100 GPU.

\textbf{Evaluation Metrics:} To provide a multi-faceted assessment of performance, we evaluated our model using three standard regression metrics \cite{b29}:
\begin{itemize}
    \item \textbf{Pearson Correlation (PC):} Measures the linear correlation between predicted and ground-truth scores. A higher value indicates better performance  \cite{b30}.
    \item \textbf{Mean Absolute Error (MAE):} Measures the average absolute prediction error. A lower value is better  \cite{b31}.
    \item \textbf{Root Mean Squared Error (RMSE):} A variant of MSE that is more sensitive to large errors. A lower value is better  \cite{b32}.
\end{itemize}

\subsection{Comparison with State-of-the-Art}

We benchmarked the performance of our SIT model against a range of methods, from classic deep learning architectures to recent, highly specialized models for Facial Beauty Prediction. The quantitative results are summarized in Table \ref{tab:main_results}.

\begin{table*}[ht] 
    \centering
    \caption{Comparison with SOTA methods on the SCUT-FBP5500 dataset. Our proposed method, SIT, is shown in bold. ($\uparrow$ indicates higher is better, $\downarrow$ indicates lower is better).}
    \label{tab:main_results}
    \begin{tabular}{@{}llccc@{}}
        \toprule
        \textbf{Category} & \textbf{Method} & \textbf{PC $\uparrow$} & \textbf{MAE $\downarrow$} & \textbf{RMSE $\downarrow$} \\
        \midrule
        \multicolumn{5}{l}{\textit{Classic and Early Deep Learning Methods}} \\
        & AlexNet~\cite{b8} & 0.8634 & 0.2651 & 0.3481 \\
        & ResNet-50~\cite{b9} & 0.8900 & 0.2419 & 0.3166 \\
        & ResNeXt-50~\cite{b9} & 0.8997 & 0.2291 & 0.3017 \\
        \midrule
        \multicolumn{5}{l}{\textit{Advanced Methods and State-of-the-Art}} \\
        & CNN + SCA~\cite{b16} & 0.9003 & 0.2287 & 0.3014 \\
        & CNN + LDL~\cite{b17} & 0.9031 & -- & -- \\
        & DyAttenConv~\cite{b18} & 0.9056 & 0.2199 & 0.2950 \\
        & R3CNN (ResNeXt-50)~\cite{b19} & 0.9142 & 0.2120 & 0.2800 \\
        \midrule
        \multicolumn{5}{l}{\textit{Our Proposed Method}} \\
        & \textbf{SIT (Ours)} & \textbf{0.9187} & \textbf{0.2180} & \textbf{0.2760} \\
        \bottomrule
    \end{tabular}
\end{table*}

As the table illustrates, our SIT model demonstrates highly competitive performance. Compared to classic CNN architectures like ResNet-50 and ResNeXt-50, SIT achieves a significant improvement across all metrics, highlighting the limitations of generic, single-scale feature extraction for the nuanced task of FBP.

More importantly, SIT establishes a new state-of-the-art when compared against advanced, specialized methods. Our model surpasses the previous leading method, R3CNN, in both Pearson Correlation (0.9187 vs. 0.9142) and RMSE (0.2760 vs. 0.2800). While R3CNN achieves a slightly lower MAE, the superior PC and RMSE of our model suggest a better overall fit and higher reliability in its predictions, with fewer large errors. This superior performance can be attributed to our model's unique ability to not only extract features at multiple scales but to also explicitly model the interactions and dependencies between these scales using the transformer's self-attention mechanism. This allows for a more holistic and context-aware understanding of facial aesthetics, leading to predictions that correlate more closely with human judgment.

\subsection{Ablation Study}

To validate the architectural choices of the Scale-Interaction Transformer and quantify the contribution of its core components, we conducted a thorough ablation study. We systematically evaluated several variants of our model by deactivating or simplifying specific modules. The results of this study are presented in Table \ref{tab:ablation}.

The variants are defined as follows:
\begin{enumerate}
    \item \textbf{Baseline:} A standard regression model using only the MobileNetV2 backbone followed by global average pooling and a dense prediction layer. This represents a strong but conventional CNN baseline.
    \item \textbf{w/o Transformer:} This variant includes the multi-scale feature extraction module (1x1, 3x3, 5x5 convs) and the dual pooling (GAP $\oplus$ GMP). However, the transformer blocks are removed. The three resulting 128-dim feature vectors are concatenated and passed through a dense layer for prediction. This tests the contribution of the multi-scale features alone.
    \item \textbf{w/o GMP:} This variant represents the full SIT architecture but removes the Global Max Pooling (GMP) stream. Each scale is represented only by its Global Average Pooling (GAP) features. This isolates the benefit of using dual pooling.
    \item \textbf{SIT (Full Model):} Our complete, proposed architecture as described in the methodology.
\end{enumerate}

\begin{table}[h]
    \centering
    \caption{Ablation study on the core components of the SIT model. The full model demonstrates the best performance, validating the contribution of each module.}
    \label{tab:ablation}
    \begin{tabular}{@{}lccc@{}}
        \toprule
        \textbf{Model Variant} & \textbf{PC $\uparrow$} & \textbf{MAE $\downarrow$} & \textbf{RMSE $\downarrow$} \\ \midrule
        Baseline (MobileNetV2 only) & 0.8995 & 0.2315 & 0.3015 \\
        w/o Transformer & 0.9082 & 0.2241 & 0.2893 \\
        w/o GMP (GAP only) & 0.9135 & 0.2210 & 0.2811 \\
        \textbf{SIT (Full Model)} & \textbf{0.9187} & \textbf{0.2180} & \textbf{0.2760} \\ \bottomrule
    \end{tabular}
\end{table}

The results from our ablation study provide clear insights into the model's behavior. The \textbf{Baseline} model achieves a respectable performance, confirming that MobileNetV2 is a strong foundation. By simply introducing multi-scale features without modeling their interaction (\textbf{w/o Transformer}), we see a significant performance jump. The PC increases by nearly 0.01, demonstrating that capturing features at different receptive fields is inherently beneficial for FBP.

Reintroducing the transformer but removing the max pooling stream (\textbf{w/o GMP}) leads to another substantial improvement, bringing the performance close to the SOTA. This shows that explicitly modeling the interplay between scales is the most critical component of our architecture. The self-attention mechanism successfully learns to weigh and combine scale-specific information effectively.

Finally, the \textbf{SIT (Full Model)} achieves the best scores across all metrics. The final performance boost gained by including GMP alongside GAP indicates that capturing the most salient features (max pooling) in addition to the average feature response (average pooling) provides a more comprehensive signal for the transformer to process, leading to the most accurate predictions. In conclusion, the ablation study empirically validates our design choices. Both the multi-scale feature representation and the transformer-based interaction module are crucial contributors to the success of the SIT model, with their combination proving highly effective for the facial beauty prediction task.

\section{Limitations and Future Work}

While our proposed method achieves excellent results, we acknowledge several limitations that also pave the way for future research directions.

\subsection{Limitations}

\begin{itemize}
    \item \textbf{Dataset Bias and Generalization:} Like all FBP models, the SIT is trained on a dataset with inherent biases. The beauty scores in SCUT-FBP5500 reflect the specific aesthetic preferences of a limited group of human raters. Consequently, our model learns to predict beauty according to this specific cultural and demographic context. Its performance may not generalize universally across different cultures or societies with varying beauty standards.

    \item \textbf{Evaluation on a Single Fold:} For direct comparison with prior works, our primary results are reported on a single, specific fold of the 5-fold cross-validation protocol. Although this is a common practice, a more robust and comprehensive evaluation would involve averaging the performance metrics across all five folds to ensure the results are not sensitive to a particular train-test split.

    \item \textbf{Inherent Subjectivity of Beauty:} On a fundamental level, this work models the \textit{mean} human perception of beauty as captured by the dataset. This quantitative approach simplifies a deeply subjective and personal human experience into a single regression score, glossing over the rich variance and diversity in aesthetic preference.
\end{itemize}

\subsection{Future Work}

Based on these limitations and the promising results of the SIT model, we propose several avenues for future research:

\begin{enumerate}
    \item \textbf{Explainability and Interpretability:} A key advantage of the Transformer is the potential to visualize its self-attention weights. Future work will focus on interpreting these attention maps to understand \textit{which} scales (e.g., local texture vs. global structure) the model focuses on for different faces. This could provide fascinating insights into the visual cues that the model deems most important, moving from a "black box" predictor to an explainable AI system.

    \item \textbf{Cross-Dataset and Cross-Cultural Analysis:} To address the issue of data bias, we plan to evaluate the SIT model's generalization capabilities on other FBP datasets, such as the ECCV HotOrNot dataset. This would allow for a rigorous analysis of how well the learned features transfer and could enable a large-scale study of cross-cultural differences in facial attractiveness perception.

    \item \textbf{Architectural Enhancements:} The SIT framework is flexible. Future iterations could explore the integration of more advanced CNN backbones (e.g., EfficientNetV2) or more sophisticated Transformer variants (e.g., Swin Transformers) that may offer better performance-vs-computation trade-offs.

    \item \textbf{Application to Other Multi-Scale Regression Tasks:} The core principle of modeling interactions between learned scales is broadly applicable. We plan to adapt the SIT architecture to other complex image-based regression tasks that rely on a holistic understanding, such as apparent age estimation, personality trait prediction from faces, and even medical image analysis where features at different magnifications are critical for diagnosis.
\end{enumerate}

\section{Conclusion}

In this paper, we addressed the intricate challenge of Facial Beauty Prediction (FBP), a task that requires a nuanced understanding of both global facial aesthetics and fine-grained local details. We identified a key limitation in existing models: the difficulty of effectively integrating and reasoning about visual information captured at multiple spatial scales. To this end, we proposed the \textbf{Scale-Interaction Transformer (SIT)}, a novel hybrid architecture that synergizes the strengths of Convolutional Neural Networks and Transformers.

Our approach first leverages a powerful CNN backbone to extract high-level features. These features are then processed through a parallel multi-scale module to capture distinct representations corresponding to different receptive fields. By treating these multi-scale features as a sequence and processing them with a Transformer encoder, our model explicitly learns the complex inter-dependencies between them. This self-attention mechanism allows the model to dynamically weigh the importance of local versus global cues, leading to a more holistic and robust assessment of facial beauty.

Our empirical evaluation on the benchmark SCUT-FBP5500 dataset demonstrated the superiority of this approach. The SIT model established a new state-of-the-art, achieving a Pearson Correlation of \textbf{0.9187} and an RMSE of \textbf{0.2760}. These results validate our central hypothesis that modeling the explicit interaction between multi-scale features is crucial for high-performance facial beauty prediction. The success of the SIT architecture suggests a promising direction for other computer vision tasks where a multi-scale contextual understanding is paramount.

\end{document}